\pdfoutput=1

\documentclass[11pt]{article}

\usepackage{acl}

\usepackage{times}
\usepackage{latexsym}
\usepackage{fullpage}
\usepackage{amsfonts}
\usepackage{amsmath, amsthm, amssymb}
\usepackage{enumitem}
\usepackage{mathtools}
\usepackage{multicol}
\usepackage{booktabs,tabularx}

\usepackage[T1]{fontenc}

\usepackage[utf8]{inputenc}

\usepackage{microtype}

\title{Quantitative Stopword Generation for Sentiment Analysis via Recursive and Iterative Deletion}

\author{Daniel M. DiPietro \\
  \texttt{dipietrodaniel131@gmail.com}\\}

\begin{document}
\maketitle
\begin{abstract}
Stopwords carry little semantic information and are often removed from text data to reduce dataset size and improve machine learning model performance. Consequently, researchers have sought to develop techniques for generating effective stopword sets. Previous approaches have ranged from qualitative techniques relying upon linguistic experts, to statistical approaches that extract word importance using correlations or frequency-dependent metrics computed on a corpus. We present a novel quantitative approach that employs iterative and recursive feature deletion algorithms to see which words can be deleted from a pre-trained transformer's vocabulary with the least degradation to its performance, specifically for the task of sentiment analysis. Empirically, stopword lists generated via this approach drastically reduce dataset size while negligibly impacting model performance, in one such example shrinking the corpus by $28.4\%$ while improving the accuracy of a trained logistic regression model by $0.25\%$. In another instance, the corpus was shrunk by $63.7\%$ with a $2.8\%$ decrease in accuracy. These promising results indicate that our approach can generate highly effective stopword sets for specific NLP tasks.
\end{abstract}

\section{Introduction}

Stopwords are words that carry little semantic information and, as a result, are often omitted when performing natural language processing (NLP) tasks. Originally used to accelerate document retrieval and search engine algorithms, stopwords are now commonly applied to NLP tasks involving machine learning, such as model training \cite{1}. Removing stopwords can reduce dataset size, which can lead to faster training on less sophisticated hardware; removing stopwords can also improve model performance \cite{2, 3}.

Often, stopwords are generated via qualitative, intuitive approaches. Researchers have also employed model-agnostic statistical approaches that generate specific stopword sets for a given corpus of text. One early, naive approach involves ranking all words in a given corpus by frequency and simply omitting the top-$n$ most frequent words \cite{1}. However, more sophisticated approaches have developed over time. 

\citet{4} developed a novel statistical technique for stopword generation uses Chi-squared statistics to generate statistical correlations between each word in a corpus and each document category. These correlations can be further normalized based on each word's frequency and used to generate a corpus-specific list of stopwords. In practice, these stopword sets decreased dataset size and often improved performance.

\citet{5} employed a different approach, using the product of term frequency and document frequency to form an ``importance score'' for each word in the vocabulary; this was based on the idea that unimportant words appear not only very frequently within documents, but also frequently across documents. Based on these importance scores, they generated stopword lists based on a user-specified tolerance; these lists closely matched lists generated by linguistic experts.

Finally, \citet{6} used probabilistic metrics for stopword identification. They computed how likely each word is to appear in document in the corpus, conditioned on the document being a specific category. Then, based on the distribution of words in the corpus at large, they computed how likely the word is to appear across the corpus in general, positing that the most semantically important words are more likely to appear in specific, relevant classes of documents than they are to appear across the corpus in general. Using this approach, they ranked each word and formed stopword sets; documents in the same class clustered together similarly with and without stopwords removed.

Thus far, all approaches discussed have relied solely upon ``importance'' metrics extracted from a corpus and assigned to each word. However, modern machine learning models often operate in unclear, complicated ways. In some cases, they demonstrate an incredible understanding of language. Specifically, transformers often know which words are important and which are not for their specific NLP task via their attention mechanism \cite{transormers}. Here, we present a novel quantitative method for stopword generation that exploits the performance of large-scale language models by using them as a tool for stopword generation. Specifically, we employ two feature selection algorithms, iterative deletion and recursive deletion, to see which stopwords can be removed from a pre-trained DistilBERT transformer with the least degradation to its area under the curve (AUC) performance on the SST2 sentiment benchmark. We use these degradations as a proxy for word importance, crafting several stopword lists. We then test the effectiveness of these stopword lists by using them to train a sentiment analysis logistic regression model.

Our results indicate that stopwords lists generated via this approach can be used to drastically reduce corpus size without hindering the performance of trained models. Thus, our techniques can lead to smaller datasets that reduce the computational burden of training machine learning models.

\begin{figure*}
    \centering
    \includegraphics[width=\linewidth]{"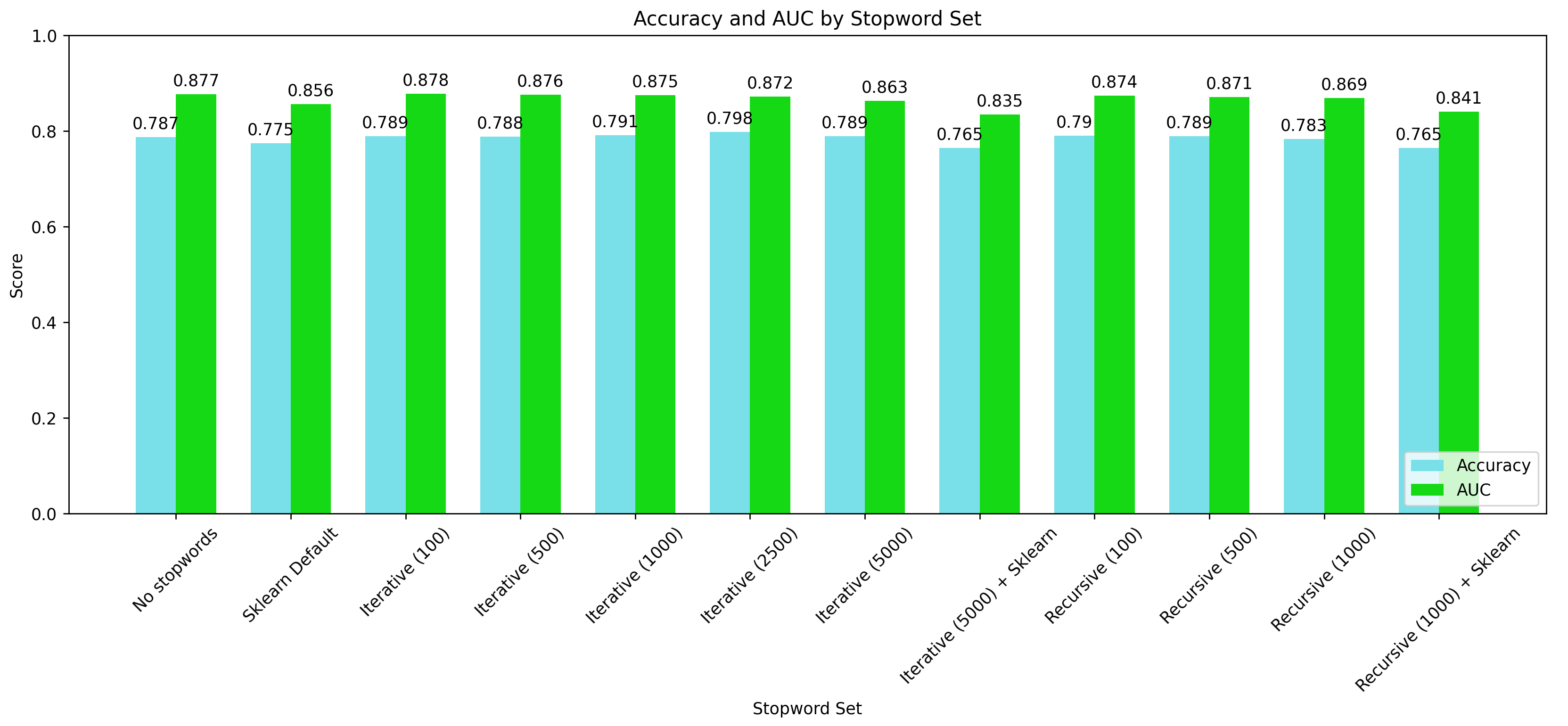"}
    \caption{Logistic Regression (vectorized with TFIDF unigrams) performance when trained on SST2 train with a given set of stopwords removed. Each logistic regression model was then benchmarked on the SST2 validation dataset, whereas the stopwords were generated from the SST2 test dataset.}
    \label{fig:metrics}
\end{figure*}

\section{Methodology}

To determine the importance of each word for the task of sentiment analysis, we begin with a pre-trained sentiment analysis transformer model. Then, we remove words from the vocabulary of its test dataset using recursive and iterative approaches, measuring the degradation of overall AUC as a proxy for word importance. Stopword lists of length $n$ have the $n$ least important words.

\subsection{Pre-Trained Language Models and SST2}

Our chosen transformer is a DistilBERT model that has been pre-trained on the Stanford Sentiment Treebank v2 (SST2) training dataset \cite{distilbert}. We pair it with a DistilBERT auto-tokenizer, also pre-trained on SST2. DistilBERT models offer impressive performance while being computationally efficient, which is important for our task given its computational complexity. SST2 is a sentiment analysis dataset widely used as a performance benchmark and freely available from dozens of online repositories \cite{sst2}. Its training, test, and validation sets have $6920$, $1821$, and $872$ examples respectively.

\subsection{Generating a Vocabulary}

To construct a vocabulary of unique tokens to rank, we remove punctuation from the SST2 test dataset and split on spaces. We opt not to use the DistilBERT auto-tokenizer for this task due to its uninterpretability. By defining the vocabulary in this way, the tokens are clear, interpretable English words that can be applied to any dataset and model.

\subsection{Iterative Deletion}

Suppose we have a corpus $\mathbf{X}$ and vocabulary $\mathbf{V}=\{v_1, \dots, v_n\}$. Let $\mathbf{X} \bigoplus v_i$ denote the operation of removing token $v_i$ from the corpus $\mathbf{X}$.

Suppose we have a model $f$. Denote the AUC metric of model $f$ on corpus $X$ as $\text{AUC}(f, \mathbf{X})$.

To perform an iterative deletion, we generate the values $\text{AUC}(f, \mathbf{X} \bigoplus v_1) - \text{AUC}(f, \mathbf{X})$ for all $v_i \in \mathbf{V}$, which is equal to the AUC degradation of model $f$ on corpus $\mathbf{X}$ when token $v_i$ is removed. This metric can be used to measure word importance for the task of sentiment analysis.

\subsection{Recursive Deletion}

Recursive deletions function somewhat similarly to iterative deletions. First, we perform an iterative deletion with starting corpus $\mathbf{X}$ and vocabulary $\mathbf{V}$. Suppose the least important word obtain from this deletion is $v_{n_1}$. We denote $v_{n_1}$ as our first stopword. Then, we perform another iterative deletion with starting corpus $\mathbf{X} \bigoplus v_{n_1}$ and vocabulary $\mathbf{V} \backslash \{v_{n_1}\}$; the least important word of this deletion is the second stopword. Generally, to obtain the $i$-th stopword, we select the least important word from an iterative deletion with corpus $\mathbf{X} \bigoplus v_{n_1} \bigoplus \dots \bigoplus v_{n_{i-1}}$ and vocabulary $\mathbf{V} \backslash \{v_{n_1}, \dots, v_{n_{i-1}}\}$.

Recursive deletions assess how the importance of words changes based on what words have been previously deleted, which is a useful characteristic for stopword list generation. However, they are computationally expensive, requiring an iterative deletion to select a single stopword.

\subsection{AUC Degradation as an Importance Metric}

We use area under the curve (AUC) degradation as our importance metric as AUC is calculated based on label prediction probability rather than actual label output of a model. Often, removing words may not necessarily change the output label of a model, but it may lead to a decrease in prediction probability performance. Thus, AUC as a metric captures a holistic view of word importance compared to label-based metrics like accuracy or F1-score.

\section{Results}

Using iterative deletion with DistilBERT as defined above, we assigned word importances to the $6862$ unique tokens found in SST2 test. The $10$ least important tokens were $\{$gets, artsy, having, just, under, sulky, bowel, think, before, lrb$\}$. Other notable tokens in the worst $25$ were ``affronted'' and ``disappointed.'' The $10$ most important tokens were $\{$not, but, n't, never, is, no, without, too, and, a$\}$. Other notable important tokens are ``in'' and ``of'' as the 14th and 21st most important tokens respectively.

Using recursive deletion, we generated a list of the $1000$ least important tokens in SST2, due to computational cost. The $10$ least important tokens were $\{$gets, artsy, having, sulky, just, bowel, think, before, skip, lrb$\}$. As we only ranked the $1000$ least important words, there is no importance list.

Using our iterative and recursive word importance metrics, we generated several stopword lists of varying size and composition (in some cases, also incorporating the Sklearn default list of stopwords). Then, we trained a logistic regression model (vectorized with unigrams and TFIDF) on SST2 train with each stopword set omitted. We benchmarked the performance of these models on SST2 validation; note that, since we used SST2 test to generate our stopwords, there are no issues with bias. These results are depicted in Figure \ref{fig:metrics}.

\begin{figure*}
    \centering
    \includegraphics[width=\linewidth]{"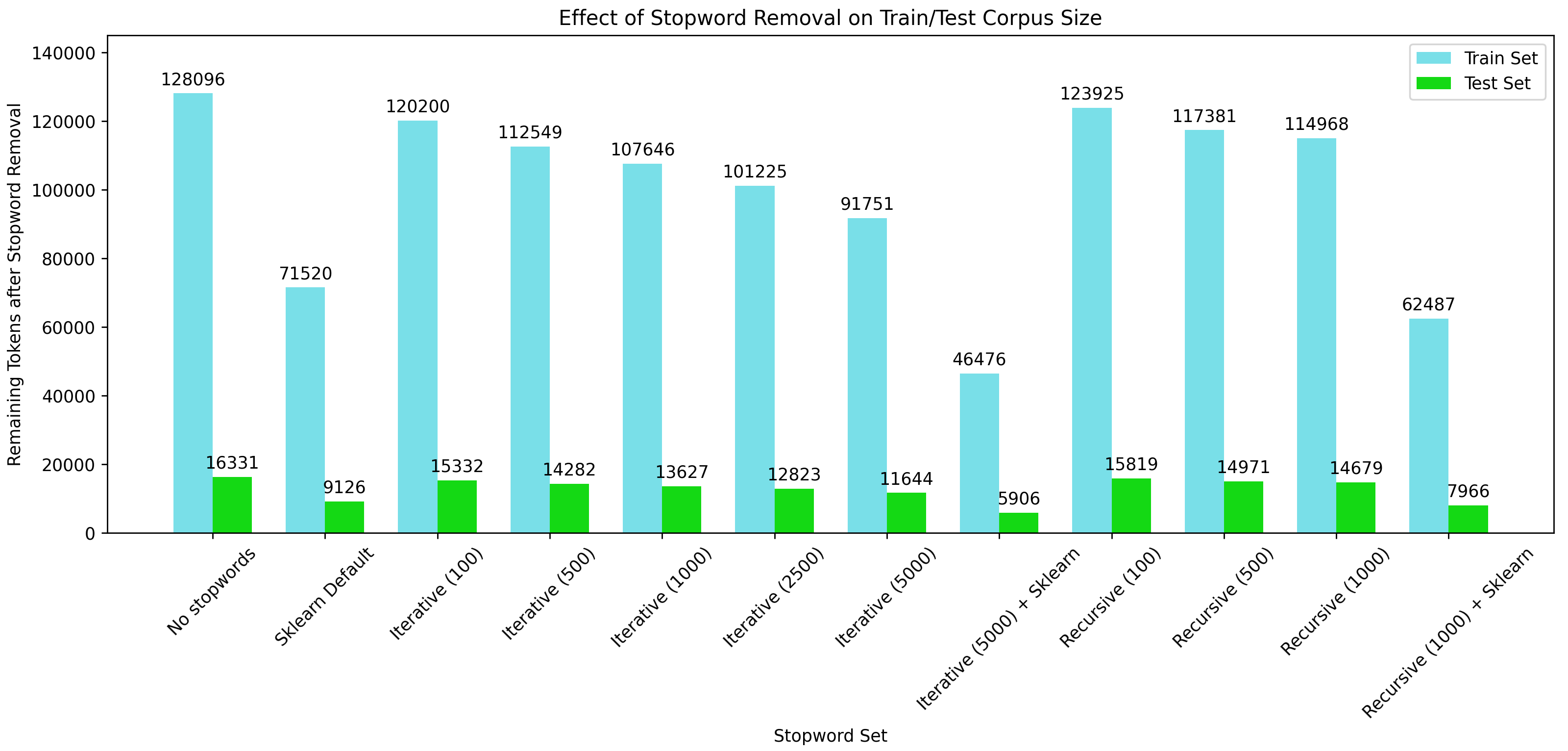"}
    \caption{Dataset reduction to the logistic regression train (SST2 train) and test (SST2 val) datasets when applying different sets of stopwords.}
    \label{fig:reduction}
\end{figure*}

Additionally, as an important consequence of stopword sets is their ability to reduce dataset size, we report the subsequent reduction in dataset size for our training and testing datasets (SST2 test and SST2 validation respectively) with each stopword set removed in Figure \ref{fig:reduction}.

\section{Discussion}

Our stopword sets seem to perform quite well, reducing corpus size while minimally impacting model performance. For example, the Recursive (1000) + Sklearn stopword cut the net corpus size in half while only degrading accuracy by $0.022$. In some cases, stopword sets even improved accuracy, such as the Iterative (1000) stopword set, which increased accuracy from $0.787$ to $0.791$ while reducing corpus size by $16\%$.

In practice, the iterative stopword sets often outperformed their recursive counterparts both in dataset reduction and model performance (the $1000$ size recursive set is an exception). As recursive stopword sets are measuring the impact of word removal on the importance of remaining words, it's possible that the relationships between words found in the SST2 test dataset were overfitted to in the stopword set and don't map well to SST2 val. 

Of particular note is that many of our stopwords are not what would commonly be considered stopwords. Words like ``artsy'' and ``disappointed'' are among the least important words, even though they're somewhat nuanced, highly specific, and seem to convey some sentiment information, at least to us. However, large-scale language models work in complicated, blackbox-esque ways, and, ultimately, it's possible that these words just aren't that important for a sentiment task in the way that a transformer learns it. Indeed, their appearance may be correlated with other words or certain syntactic structures that offer better sentiment information.

Relatedly, many common stopwords, rather than appearing among the least important words, actually appear among the most important words: some examples are ``is'', ``a'', ``in'', and ``of.'' This may seem counterintuitive, but it's important to note that our DistilBERT model was not trained with stopwords. Even though these words don't offer sentiment information, it's possible that they're essential for providing the language syntax that the transformer was trained to understand, which is why they may not be picked up by our approach.

Our work offers no major ethical issues: none of the stopwords are associated with protected classes, and SST2 is a widely used and thoroughly vetted dataset. However, our approach requires large sentiment analysis datasets, which limits it to high-resource languages, and the carbon implications of recursive deletion are notable, as these searches took approximately 48 hours of GPU run-time.

\section{Conclusion}

Our approach offers a highly effective technique for generating stopword sets for specific NLP tasks, which can reduce dataset size and allow for training models on cheaper hardware. In the future, we can perform this process on larger datasets, with more sophisticated transformer models, and for different NLP tasks. Additionally, we can perform more sophisticated linguistic analyses to better understand why some words were unintuitivately designated or not designated as stopwords.

\bibliography{custom}

\begin{thebibliography}{9}
\expandafter\ifx\csname natexlab\endcsname\relax\def\natexlab#1{#1}\fi

\bibitem[{Dolamic and Savoy(2010)}]{3}
Ljiljana Dolamic and Jacques Savoy. 2010.
\newblock When stopword lists make the difference.
\newblock \emph{Journal of the American Society for Information Science and
  Technology}, 61(1):200--203.

\bibitem[{Ferilli(2021)}]{5}
Stefano Ferilli. 2021.
\newblock Automatic multilingual stopwords identification from very small
  corpora.
\newblock \emph{Electronics}, 10(17):2169.

\bibitem[{Hao and Hao(2008)}]{4}
Lili Hao and Lizhu Hao. 2008.
\newblock Automatic identification of stop words in chinese text
  classification.
\newblock In \emph{2008 International conference on computer science and
  software engineering}, volume~1, pages 718--722. IEEE.

\bibitem[{Kaur and Buttar(2018)}]{2}
Jashanjot Kaur and P~Kaur Buttar. 2018.
\newblock A systematic review on stopword removal algorithms.
\newblock \emph{International Journal on Future Revolution in Computer Science
  \& Communication Engineering}, 4(4):207--210.

\bibitem[{Sanh et~al.(2019)Sanh, Debut, Chaumond, and Wolf}]{distilbert}
Victor Sanh, Lysandre Debut, Julien Chaumond, and Thomas Wolf. 2019.
\newblock Distilbert, a distilled version of bert: smaller, faster, cheaper and
  lighter.
\newblock \emph{arXiv preprint arXiv:1910.01108}.

\bibitem[{Sch{\"u}tze et~al.(2008)Sch{\"u}tze, Manning, and Raghavan}]{1}
Hinrich Sch{\"u}tze, Christopher~D Manning, and Prabhakar Raghavan. 2008.
\newblock \emph{Introduction to information retrieval}, volume~39.
\newblock Cambridge University Press Cambridge.

\bibitem[{Socher et~al.(2013)Socher, Perelygin, Wu, Chuang, Manning, Ng, and
  Potts}]{sst2}
Richard Socher, Alex Perelygin, Jean Wu, Jason Chuang, Christopher~D Manning,
  Andrew~Y Ng, and Christopher Potts. 2013.
\newblock Recursive deep models for semantic compositionality over a sentiment
  treebank.
\newblock In \emph{Proceedings of the 2013 conference on empirical methods in
  natural language processing}, pages 1631--1642.

\bibitem[{Vaswani et~al.(2017)Vaswani, Shazeer, Parmar, Uszkoreit, Jones,
  Gomez, Kaiser, and Polosukhin}]{transormers}
Ashish Vaswani, Noam Shazeer, Niki Parmar, Jakob Uszkoreit, Llion Jones,
  Aidan~N Gomez, {\L}ukasz Kaiser, and Illia Polosukhin. 2017.
\newblock Attention is all you need.
\newblock \emph{Advances in neural information processing systems}, 30.

\bibitem[{Wilbur and Sirotkin(1992)}]{6}
W~John Wilbur and Karl Sirotkin. 1992.
\newblock The automatic identification of stop words.
\newblock \emph{Journal of information science}, 18(1):45--55.

\end{thebibliography}
\bibliographystyle{acl_natbib}

\end{document}